# LLMs vs. Traditional Sentiment Tools in Psychology: An Evaluation on Belgian-Dutch Narratives

**Ratna Kandala, Katie Hoemann**

**Department of Psychology, University of Kansas, USA**

**Abstract**

Understanding emotional nuances in everyday language is crucial for computational linguistics and emotion research. While traditional lexicon-based tools like LIWC and Pattern have served as foundational instruments, Large Language Models (LLMs) promise enhanced context understanding. We evaluated three Dutch-specific LLMs (ChocoLlama-8B-Instruct, Reynaerde-7B-chat, and GEITje-7B-ultra) against LIWC (27) and Pattern (9) for valence prediction in Flemish, a low-resource language variant. Our dataset comprised approximately 25,000 spontaneous textual responses from 102 Dutch-speaking participants, each providing narratives about their current experiences with self-assessed valence ratings (–50 to +50). Surprisingly, despite architectural advancements, the Dutch-tuned LLMs underperformed compared to traditional methods, with Pattern showing superior performance. These findings challenge assumptions about LLM superiority in sentiment analysis tasks and highlight the complexity of capturing emotional valence in spontaneous, real-world narratives. Our results underscore the need for developing culturally and linguistically tailored evaluation frameworks for low-resource language variants, while questioning whether current LLM fine-tuning approaches adequately address the nuanced emotional expressions found in everyday language use.

## 1 Introduction

The increasing use of Large Language Models (LLMs) in everyday applications has raised important questions about their ability to understand emotional nuances, particularly as users seek empathetic interactions with AI systems. Current LLMs lack psychological mechanisms for genuine empathy (35), making culturally and linguistically tailored valence detection critical, especially for low-resource languages like Flemish (Belgian-Dutch). Traditional lexicon-based tools like LIWC (29) and Pattern (9) have dominated valence analysis through predefined word-emotion mappings. While LIWC excels with validated dictionary categories and Pattern offers robust rule-based analysis, both struggle with spontaneous, context-rich narratives. For example, LIWC would misclassify sarcastic expressions like "This amazing Monday has me stuck in traffic for hours - what a joy!" as positive, failing to detect contextual frustration. Most sentiment analysis research focuses on high-resource languages like English (16), while Flemish remains underrepresented despite being spoken by 6.5 million Belgians (20). Flemish exhibits distinct features from standard Dutch, including regional vocabulary ("fuif" for "party"), divergent pronouns ("gij" vs. "jij"), and multilingual influences, challenging existing tools. This study addresses three critical gaps: linguistic inequity in NLP research, ecological validity (moving beyond constrained social media data (33; 34)), and methodological rigor using self-reported rather than annotator-based ground truth. We compare lexicon-based tools (LIWC, Pattern) with Dutch-tuned LLMs (ChocoLlama-8B-Instruct, Reynaerde-7B-chat, GEITje-7B-ultra) using nearly 25,000 open-ended Flemish narratives from 102 participants collected via ambulatory assessment. Participants provided real-time experiences with self-rated valence scores (–50 to +50). Our findings reveal that state-of-the-art Dutch LLMs underperform lexicon-based methods, underscoring needs for linguistically tailored models and equitable AI development that respects linguistic diversity in applications from mental health interventions to social science research.

## 2 Prior Work

Sentiment analysis, detecting emotional valence in language, is widely applied across healthcare, customer service, and misinformation detection, with traditional lexicon-based tools like LIWC (29) and Pattern (9) dominating psychological research due to their interpretability, though they lack contextual awareness and struggle with sarcasm and domain-specific cues (22). While recent studies highlight LLMs' potential for English valence detection (19), their performance in low-resource languages remains understudied, with untuned models often failing to capture emotion intensity due to English-centric training data (18; 16). For Flemish, spoken by 6.5 million Belgians with distinct linguistic features, sentiment analysis research is sparse, with existing work like Reusens (33) limited to social media data that distorts emotional expression through character limits and platform constraints (34). We address these gaps by evaluating three Dutch-tuned LLMs against LIWC and Pattern on nearly 25,000 open-ended Flemish narratives with self-reported valence scores, using ambulatory assessment data that preserves linguistic authenticity and provides ecologically valid ground truth compared to external annotator approaches (26; 12).

## 3 Lexicon-Based Text Analyses and LLMs Chosen

Lexicon-based methods employ dictionaries to detect sentiments, valued for their interpretability and ease of implementation. Linguistic Inquiry and Word Count (LIWC) (27), adapted for Dutch (5), maps words across 90+ categories including emotional tone (PosEmo and NegEmo). The Dutch version contains 6,614 words (1,227 PosEmo, 1,474 NegEmo), estimating valence by calculating percentage of dictionary-matched words. However, as a proprietary tool, LIWC lacks transparency and doesn't incorporate negation or sarcasm detection rules. Pattern.nl (9) is an open-source Python package providing "plug-and-play" sentiment analysis without machine learning expertise for psychology researchers. It uses a handcrafted lexicon of 3,304 Dutch lemmas (97% adjectives) with polarity values [-1,1]. The algorithm performs chunking, chunk-level scoring with intensifier/negation rules, and sentence-level aggregation. However, the lexicon, compiled from product reviews, may score affective texts as neutral when words fall outside this domain. We selected three open-weight Dutch-tuned LLMs for valence estimation: ChocoLlama-8B-Instruct, GEITje-7B-Ultra, and Reynaerde-7B-Chat, chosen for architectural diversity and data privacy considerations. ChocoLlama-8B-Instruct (20) is built on Llama-3-8B with rank-16 LoRA on 32B Dutch tokens, followed by supervised fine-tuning and DPO alignment. GEITje-7B-Ultra (38) is based on Mistral 7B with full-parameter pretraining on 10B Dutch tokens, followed by SFT and DPO alignment, reported as the best GEITje variant. Reynaerde-7B-Chat (32) uses rank-8 QLoRA adapters on Llama-2-7B-hf, fine-tuned on 400K Dutch chat pairs with DPO alignment for safe dialogue. Both GEITje and Reynaerde have outperformed earlier ChocoLlama variants on Dutch benchmarks, though their psychological research efficacy remains unvalidated (20).

## 4 Methodology

### 4.1 Data

The dataset comprises 24,854 open-ended Dutch-language text entries (medium-scale) collected longitudinally over 70 days from 102 participants (age : 18–65, $\mu = 26.47$, $\sigma = 8.87$) in Belgium. Participants were native Dutch speakers with smartphone access, recruited via online/posted flyers and referrals (most were native Belgians, a few were from the Netherlands and living in Belgium). Using a dedicated app, they received four daily prompts asking (in Dutch), "What is going on now or since the last prompt, and how do you feel about it?" Responses were either short text snippets (3–4 sentences) or 1-minute voice recordings, yielding a temporally structured dataset with potential for time-series or multimodal analysis (text + audio). For robustness, eligibility criteria ensured linguistic consistency (native speakers) and device reliability (smartphone ownership). Further methodological details, including ethical protocols, are provided in the Supplementary Material. The dataset can be made available to interested persons after acceptance (in compliance with GDPR). Nevertheless, the users' self-reported valence scores, the ratings by each of these tools/models will be made available online.

### 4.2 Prompt for the LLMs

To enable sentiment analysis on this dataset, we designed a standardized prompt to elicit valence ratings from various Dutch-language large language models (LLMs). We experimented with both Dutch and English prompts but did not observe any significant improvement in LLM performance with the Dutch prompts. Similarly, we found no notable difference between zero-shot and few-shot settings (see Supplementary Materials for details). So we proceeded ahead with the English prompt:

"You are a Dutch language expert analyzing the valence of Belgian Dutch texts. Participants responded to: 'What is going on now or since the last prompt, and how do you feel about it?' Carefully read the response of the participant: {text}. Your task is to rate its sentiment from 1 (very negative) to 7 (very positive). Return ONLY a single numerical rating enclosed in brackets, e.g. [X], with no additional text.

Output Format: [number]"

The placeholders {text} are filled with texts.

### 5 Results: Performance of LLMs versus Lexical Tools

This section presents the results of the models and their comparison with self-reported valences of the users. Coverage analysis revealed stark differences between methods, with lexicon-based tools (LIWC and Pattern) achieving near-complete coverage at 99.9% (24,848/24,852 texts), while Dutch-tuned LLMs showed substantial variability: ChocoLlama-8B-Instruct processed 69.9% of texts, GEITje-7B-ultra only 38.0%, and Reynaerde-7B-chat a mere 1.8%. These coverage gaps in LLMs risk introducing selection bias and complicate fair performance evaluations compared to the universal processing capability of lexicon tools. To assess alignment with human judgment, we evaluated how well each model's predictions correlated with users' self-reported valence ratings (ranging from –50 to +50) using both Pearson's r for linear association and polyserial correlation for ordinal-continuous relationships. Higher correlations indicated that model predictions more closely tracked the writers' own emotional assessments of their narratives. Pattern demonstrated superior lexicon-based performance with correlations of $r = 0.31$ (both Pearson and Polyserial), outperforming LIWC15-PosEmo ($r = 0.21/0.23$) and LIWC15-NegEmo ($r = -0.23/- 0.23$) across nearly all 24,852 texts. Among LLMs, ChocoLlama-8B-Instruct showed the strongest performance on its 17,378 processed texts (Pearson $r = 0.35$, Polyserial $r = 0.40$), significantly outperforming Pattern ($t = 4.02$, $p < 0.001$) on this subset. GEITje-7B-ultra achieved similar correlations ($r = 0.35/0.44$) across 9,445 texts but did not significantly differ from Pattern ($t = -1.20$, $p = 0.23$), while Reynaerde-7B-chat underperformed with correlations of only $r = 0.18/0.24$ on 446 texts, showing no significant advantage over lexicon methods. The key trade-off emerged between LLM accuracy and coverage: ChocoLlama and GEITje achieved higher correlations than lexicon tools on their processed subsets, but their limited coverage (69.9% and 38.0% respectively) versus Pattern's near-universal coverage (99.9%) raises questions about practical applicability and potential selection bias.

Table 1: Correlation and coverage metrics across sentiment models

| Model | Coverage (N / %) | Pearson $r$ | Polyserial $r$ |
|---|---|---|---|
| LIWC (posemo) | 24,848 / 99.9% | 0.21 | 0.23 |
| LIWC (negemo) | 24,848 / 99.9% | -0.23 | -0.23 |
| Pattern.nl | 24,848 / 99.9% | 0.31 | 0.31 |
| ChocoLlama-8B-Instruct | 17,378 / 69.9% | 0.35 | 0.40 |
| GEITje-7B-ultra | 9,445 / 38.0% | 0.35 | 0.44 |
| Reynaerde-7B-chat | 446 / 1.8% | 0.18 | 0.24 |

## 6 Discussion

Our evaluation of Dutch-finetuned LLMs against lexicon-based tools for Flemish valence detection revealed that while models like ChocoLlama-8B-Instruct and GEITje-7B-ultra showed moderate correlations with human ratings, they suffered from incomplete coverage and systematic domain mismatch between training data and real-world narratives. These findings challenge assumptions about LLM superiority in low-resource settings. The underperformance stems from fundamental training data limitations. Major LLMs exhibit severe English-centric bias - Llama-2's pretraining comprises 89.7% English content, while GPT-3 contains roughly 92.65% English tokens (15). Although adaptation efforts like ChocoLlama incorporated 32 billion Dutch tokens for continued pretraining (20), overreliance on formal corpora (legal documents, social media) creates models ill-suited for colloquial, first-person narratives characteristic of daily emotional expression. This domain mismatch undermines valence accuracy because LLMs lack exposure to lexical and pragmatic cues signaling emotion in everyday language (4). Conversely, LIWC and Pattern.nl, grounded in psycholinguistic principles and manually curated word lists, retain sensitivity to informal affective expressions (28; 13). Pattern.nl's superior performance likely reflects its composite polarity scoring versus LIWC's raw emotion-word counts. Addressing LLM limitations requires augmenting training data with narrative-style corpora, creating hybrid lexicon-LLM models, and developing standardized Flemish valence benchmarks (3). Until LLMs overcome domain mismatches, lexicon tools remain indispensable for scalable, ecologically valid valence analysis in low-resource languages like Flemish.

## 7 Conclusion and Limitations

This study demonstrates that lexicon-based tools like Pattern.nl remain the gold standard for valence analysis in low-resource Flemish, outperforming Dutch-tuned LLMs in coverage and reliability despite contextual limitations. Our findings highlight critical gaps in current LLM approaches: incomplete coverage, domain mismatch between formal training data and colloquial narratives, and systematic underrepresentation of Flemish linguistic features. Several limitations constrain our findings. First, Dutch and Flemish remain critically under-represented in training data and available LLMs, reflecting broader systemic gaps in low-resource language NLP (20). Our evaluation of three Dutch-tuned models does not account for larger architectures (e.g., LLaMA-2 70B) or hybrid approaches that may outperform current benchmarks. Second, while our corpus of nearly 25,000 narratives provides ecological validity, reliance on a single prompt structure risks conflating valence expression with elicitation method, limiting generalizability. Finally, lexicon tools, though reliable, lack contextual nuance for resolving sarcasm or cultural idioms and cannot replicate human empathy(25). Moving forward, establishing standardized Flemish valence benchmarks grounded in manually annotated narratives will enable targeted improvements in data augmentation, bias mitigation, and hybrid modeling. The integration of LLMs with lexicon-driven frameworks may bridge the gap between scalable automation and culturally grounded emotion research. Until such synergies are realized, lexicon tools will continue to anchor valence analysis in low-resource contexts, ensuring both methodological rigor and practical applicability for psychological research and mental health applications.

# A Supplementary Material

## A.1 Data Preparation

Participants were recruited from the community via flyers, online advertisements, and word-of-mouth referrals. Eligibility criteria included being native Dutch speakers, residing in Belgium, being at least 18 years of age, and owning a fully functional smartphone. Interested individuals completed an initial eligibility survey, with eligibility criteria verified during a subsequent online introduction session.

Participants received compensation of up to €250 for participating in a 70-day (10-week) experience sampling protocol, accompanied by biweekly online surveys. The breakdown included €0.50 per completed experience sampling prompt (maximum of 280 prompts or €140 total), €10 for each short survey at 2, 4, 6, and 8 weeks (maximum €40), and €15 for each long survey at baseline and 10 weeks (maximum €30). An additional bonus of €40 was given to participants completing at least 60 days. Payment was issued in full upon study completion. Continued participation required completion of at least 75% of prompts, with verbal responses having a minimum length of 25 words. Regular compliance checks and biweekly summary reports were provided to participants, with reminders issued as necessary.

Initially, 115 participants enrolled (age range: 18–65, $M = 27.26$, $SD = 9.86$; gender: 58 female, 56 male, 1 other). Of these, 10 voluntarily withdrew, and 3 were dismissed due to poor compliance (response rates below 50%). Thus, 102 participants completed the study (age range: 18–65, $M = 26.47$, $SD = 8.87$; 52 women, 49 men, 1 other).

Ethical approval was obtained from the KU Leuven Social and Societal Ethics Committee (SMEC), protocol G-2023-6379-R3(AMD). Data collection spanned from August 2023 to July 2024. All study materials and instructions were administered in Dutch.

### A.1.1 Procedure and Materials

Participants engaged in a 70-day experience sampling protocol via a dedicated mobile application (*m-Path*; Mestdagh et al., 2023), receiving four prompts daily at pseudorandom intervals between 9 AM and 9 PM, spaced at least one hour apart. At each prompt, participants responded to "What is going on now or since the last prompt, and how do you feel about it?" Responses were typically recorded as 1-minute voice messages but could optionally be typed (3–4 sentences). Participants then rated their current emotional valence on a slider scale ranging from −50 (very unpleasant) to +50 (very pleasant). Simultaneously, *m-Path* recorded sensor data (GPS coordinates, ambient noise, step counts, and app usage for Android devices).

All assessment tools were validated previously in Dutch-speaking samples. Participants completed additional online surveys after 2, 4, 6, and 8 weeks, and a comprehensive survey at 10 weeks, which included validation questions about recently visited locations.

Participants received biweekly compliance summary reports via email, detailing prompt completion rates, description length adequacy, modality (voice vs. text), valence ratings, linguistic content trends (using LIWC 2015 Dutch translation), and accumulated compensation. These reports also indicated whether participants earned bonuses and their overall compensation.

Voice-recorded responses were automatically transcribed using a proprietary algorithm developed by the Department of Electrical Engineering (ESAT) at KU Leuven (Tamm et al., 2024). Transcripts and typed responses were integrated into a unified data file, aligned manually with online survey data into five study intervals: days 1–14, 15–28, 29–42, 43–56, and 57–70. Missing survey data were noted where applicable.

**A.1.2 Analytic Tools and Experimental Setup**

Textual responses were analyzed using Linguistic Inquiry and Word Count (LIWC) with its Dutch lexicon, generating category-specific linguistic scores for each response. Additionally, the *Pattern.nl* sentiment analysis algorithm provided continuous valence estimates from −1 (very negative) to +1 (very positive), accounting for word context and grammatical function.

Model evaluations were performed using an NVIDIA RTX-5000 GPU and implemented in PyTorch, with models obtained via Hugging Face APIs.

**A.2 Single Participant (Pilot Study)**

We conducted a preliminary pilot study using data from one participant in our dataset to identify the best-performing models. We also compared the performances of these models using both English and Dutch prompts.

**English Prompt:**
"You are a Dutch language expert analyzing the valence of Belgian Dutch texts. Participants responded to: *'What is going on now or since the last prompt, and how do you feel about it?'*
Carefully read the response of the participant: *'{text}'*.
Your task is to rate its sentiment from 1 (very negative) to 7 (very positive).
Return **only** a single numerical rating enclosed in brackets (e.g., [X]), with no additional text, explanations, or formatting.
**Output format:** [number]. Replace 'number' with the integer score (1–7)."

**Dutch Prompt:**
"Je bent een Nederlandse taalexpert die de valentie van Belgisch Nederlandse teksten analyseert. Deelnemers reageerden op: *'Wat speelt er nu of sinds de vorige beep en hoe voel je je daarover?'*
Lees zorgvuldig het antwoord van de deelnemer: *'{text}'*.
Het is jouw taak om het sentiment te beoordelen van 1 (zeer slecht) tot 7 (zeer goed). Geef **alleen** een cijfer tussen haakjes (bijv. [X]), zonder extra tekst, uitleg of opmaak. Niet uitleggen.
**Uitvoerformaat:** [getal]. Vervang 'getal' door de gehele score (1–7)."

As illustrated in Table 1, we selected **ChocoLlama-8B-Instruct**, **GEITje-7B-Ultra**, **Reynaerde-7B-Chat**, and **Llama 2-7B**. We also ran the dataset with **LIWC15** and **Pattern.nl**. The models returned values for all texts.

Among these LLMs, when prompted in English, *ChocoLlama-8B-Instruct* achieved the highest polyserial correlation ($r = 0.55$) with user ratings, followed by *GEITje-7B-Ultra* ($r = 0.42$) and *Reynaerde-7B-Chat* ($r = 0.33$). Interestingly, *Llama 2-7B* returned no values for the English prompt but performed better than *GEITje-7B-Ultra* ($r = 0.0002$) and *Reynaerde-7B-Chat* ($r = 0.42$) when prompted in Dutch.

To maintain uniformity, we proceeded with only the English prompt for further experiments. Between *LIWC15* and *Pattern.nl*, *LIWC15* showed strong negative (*negemo*: $r = -0.54$) and positive (*posemo*: $r = 0.41$) correlations, while *Pattern.nl* showed a positive correlation with user ratings ($r = 0.44$).

Table 2: Correlation coefficients across models and prompts

| Model (variable) | Prompt | Pearson $r$ | Polyserial $r$ |
|---|---|---|---|
| LIWC15 (posemo) | – | 0.41 | – |
| LIWC15 (negemo) | – | -0.54 | – |
| Pattern.nl | – | 0.44 | – |
| ChocoLlama-8B-Instruct | English | 0.47 | 0.55 |
| ChocoLlama-8B-Instruct | Dutch | 0.18 | 0.27 |
| Llama2-7B | English | N/A | N/A |
| Llama2-7B | Dutch | 0.37 | 0.46 |
| GEITje-7B-ultra | English | 0.33 | 0.42 |
| GEITje-7B-ultra | Dutch | 0.0001 | 0.0002 |
| Reynaerde-7B-Chat | English | -0.1 | 0.33 |
| Reynaerde-7B-Chat | Dutch | 0.2 | 0.42 |

In the following sections, we present the additional experiments conducted and the results observed for the entire dataset using the English prompt:

## A.3 Distribution of Ratings (All Participants/Users, LIWC, Pattern.nl)

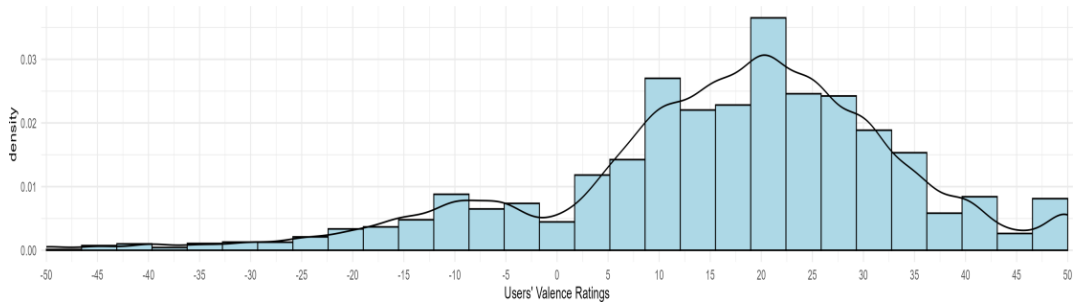

Figure 1: Distribution of users' self-reported valence ratings

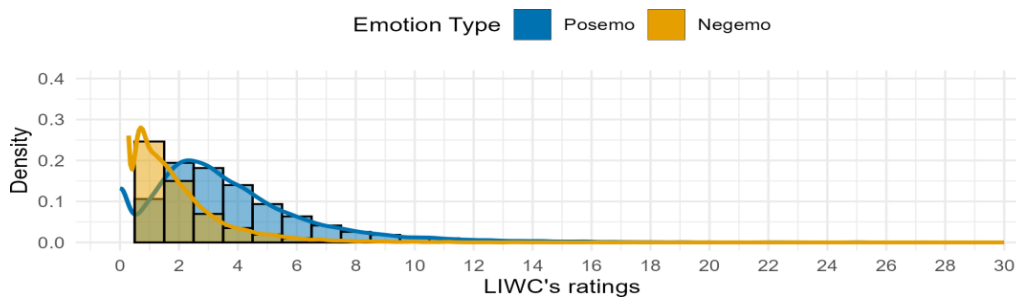

Figure 2: Distribution of LIWC's scores

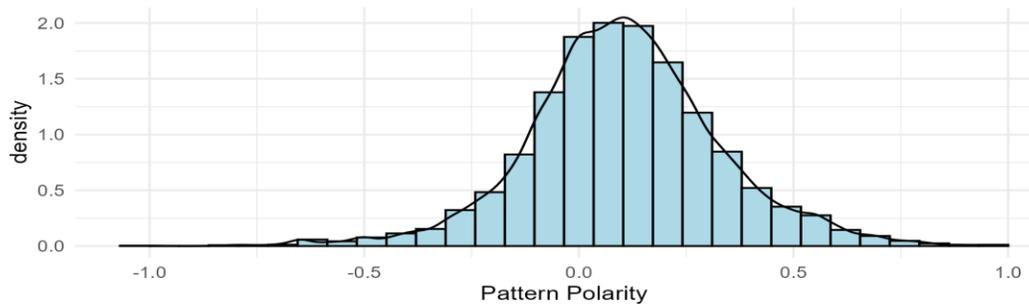

Figure 3: Distribution of Pattern.nl's polarity

## A.4 Zero-shot setting results (All participants)

In the zero-shot condition, as discussed previously in the Results section, the models did not return values for all texts, unlike LIWC and *Pattern.nl*. *ChocoLlama-8B-Instruct* returned values for 17,378 texts, *GEITje-7B-Ultra* for 9,445 texts, and *Reynaerde-7B-Chat* for only 446 estimates. *Pattern.nl* performed better on most texts than these models.

Polyserial correlation analyses revealed significant relationships between user ratings and the outputs of all three language models tested, with all user-model comparisons yielding $p < 0.001$. However, correlations between *Reynaerde* and traditional sentiment tools varied: *LIWC15 (posemo)* showed no significant association ($p = 0.12$), while *LIWC15 (negemo)* approached significance ($p < 0.05$) but did not meet the threshold for statistical reliability. In contrast, *Pattern.nl* demonstrated a modest yet statistically significant correlation with *Reynaerde* ($p = 0.022$).

*GEITje* showed strong, statistically significant correlations with all lexicon-based models—*LIWC15 (posemo)*, *LIWC15 (negemo)*, and *Pattern.nl* (all $p < 0.001$), indicating robust agreement. Similarly, *ChocoLlama*'s outputs correlated significantly with *LIWC15* and *Pattern.nl* (all $p < 0.001$).

### A.4.1 Distributional Analysis of the Models' Ratings

*ChocoLlama-8B-Instruct* showed a relatively balanced distribution compared to the other two LLMs, with peaks around 3 and 5. *GEITje-7B-Ultra*'s ratings were skewed toward scores around 5. *Reynaerde-7B-Chat*'s predictions had the narrowest distribution among the three models.

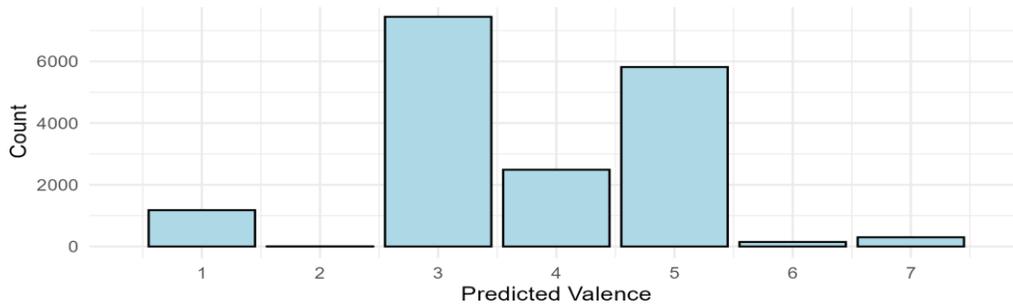

Figure 4: Distribution of Chocollama-8B-instruct's predicted valence scores

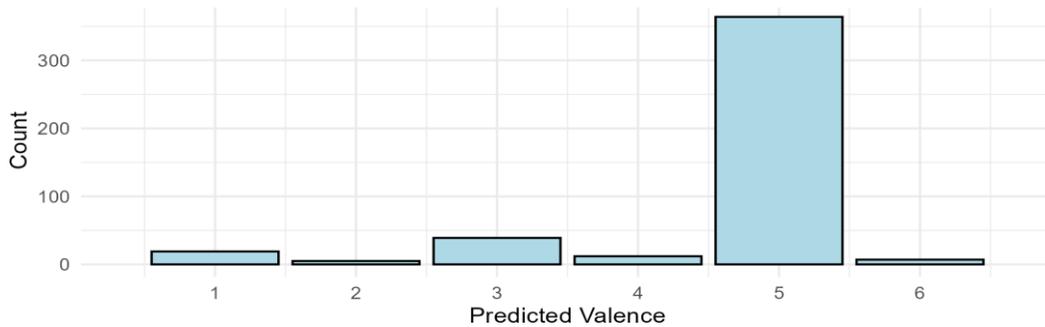

Figure 5: Distribution of Reynaerde-7B's valence scores

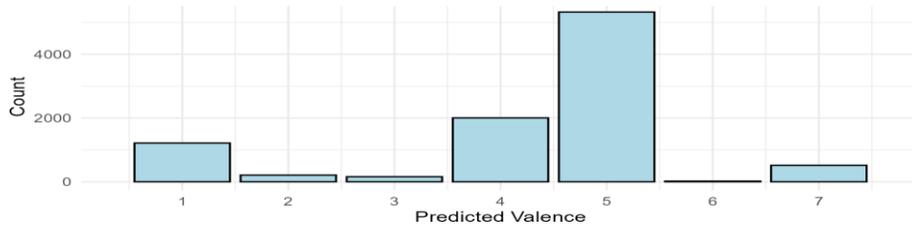

Figure 6: Distribution of GEITje-7B's valence scores

### A.4.2 Box Plot Analysis of Users' Ratings vs. Model's Predictions

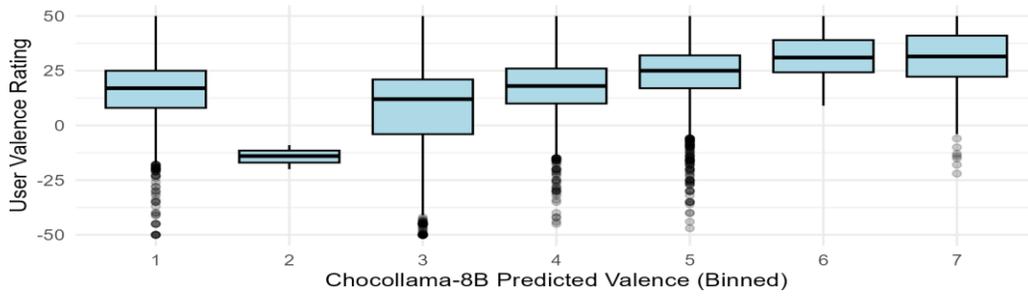

Figure 7: Chocollama-8B-instruct

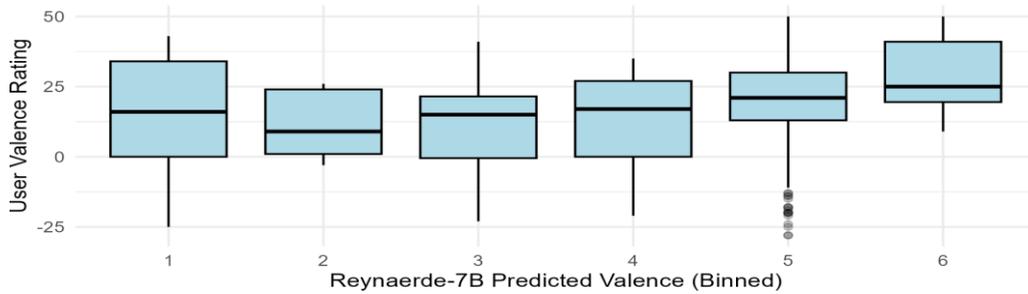

Figure 8: Reynaerde-7B

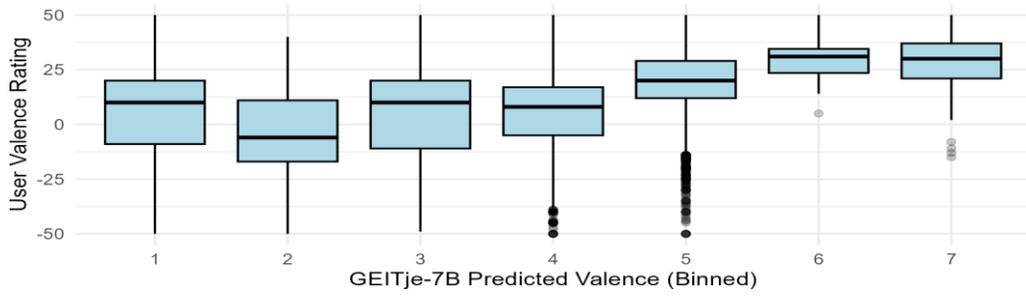

Figure 9: GEITje-7B

### A.4.3 Scatter Plots

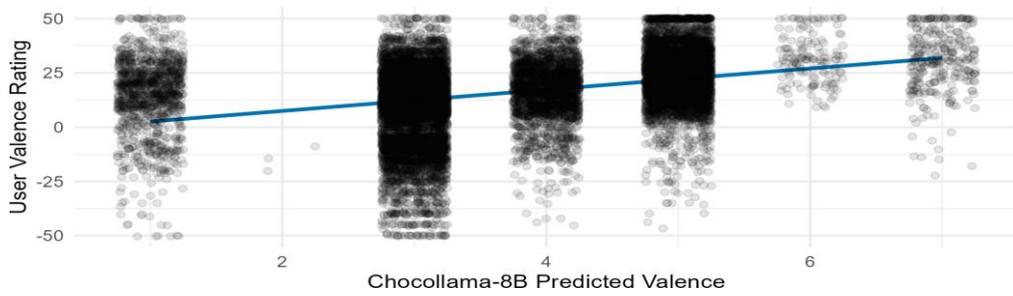

Figure 10: Chocollama-8B-instruct

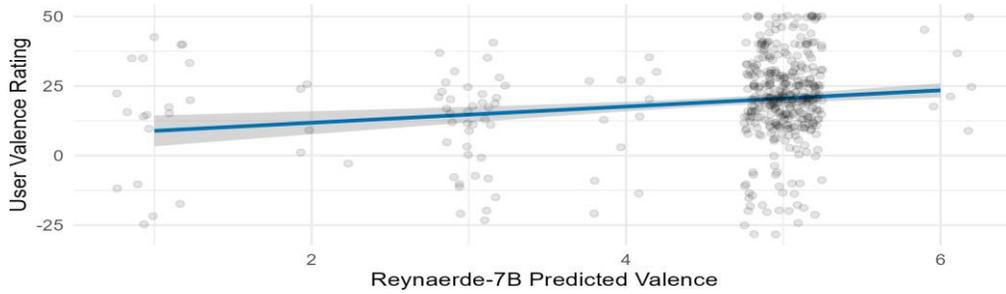

Figure 11: Reynaerde-7B

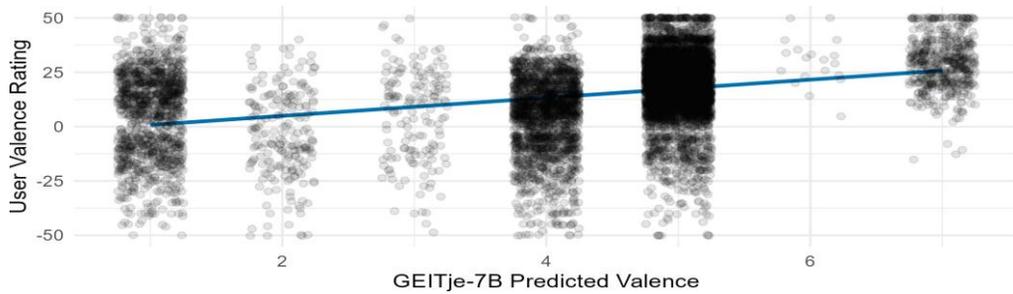

Figure 12: GEITje-7B

### A.5 Few-shot setting results (All participants)

We also attempted a few-shot setting by providing five examples from the text along with the user ratings.

#### A.5.1 Prompt used

In this prompt, "(Text)" was replaced with the actual responses of the participants.

Prompt:
"You are a Dutch language expert analyzing the valence of Belgian Dutch texts. Participants responded to the question:
*'What is going on now or since the last prompt, and how do you feel about it?'*
They also rated their own emotional valence on a continuous scale from −50 (very negative) to +50 (very positive).

Your task is to read each response (text) and rate its sentiment from 1 (very negative) to 7 (very positive). Return only a single numerical rating enclosed in square brackets, e.g., [X]. Provide no explanation or additional text.
Output format: [number].

Below are a few examples of an input and the participant's valence rating to guide your rating:

Input: (Text1)
Output: [10]

Input: (Text2)
Output: [−10]

Input: (Text3)
Output: [30]

Input: (Text4)
Output: [45]

Input: (Text5)
Output: [40]"

We observed a similar trend in the selected LLMs, which returned values for only a portion of the texts. *Pattern.nl* proved to be more effective than *LIWC*.

Table 3: Few-shot coverage and correlation coefficients across models and benchmarks wrt users

| Model | Coverage (N / %) | Pearson $r$ | Polyserial $r$ |
|---|---|---|---|
| LIWC15 (posemo) | 24,848 / 99.9% | 0.21 | 0.23 |
| LIWC15 (negemo) | 24,848 / 99.9% | -0.23 | -0.26 |
| Pattern.nl | 24,848 / 99.9% | 0.31 | 0.33 |
| Reynaerde-7B-Chat (English prompt) | 7,219 / 29.0% | 0.03 | 0.036 |
| GEITje-7B-ultra (English prompt) | 11,323 / 45.6% | -0.08 | -0.09 |
| Chocollama-8B-instruct(English prompt) | 5,266 / 21.2% | -0.03 | -0.029 |

In the few-shot setting, we provided five example texts with user ratings. Significance values (p-values) from polyserial correlation analyses were as follows:

### A.5.2 Polyserial Correlation Between the models

Table 4: Polyserial correlations and significance for model-model comparisons

| Comparison | r | p | Significance |
|---|---|---|---|
| LIWC(posemo) vs. Reynaerde-7B-Chat | 0.11 | < 0.001 | significant |
| LIWC(negemo) vs. Reynaerde-7B-Chat | −0.004 | 0.916 | non-significant |
| Pattern vs. Reynaerde-7B-Chat | 0.044 | 0.001 | significant |
| LIWC(posemo) vs. GEITje-7B-ultra | 0.068 | 0.003 | significant |
| LIWC(negemo) vs. GEITje-7B-ultra | 0.131 | < 0.001 | significant |
| Pattern vs. GEITje-7B-ultra | −0.05 | 0.700 | non-significant |
| LIWC(posemo) vs. Chocollama-8B-instruct | −0.052 | 0.479 | non-significant |
| LIWC(negemo) vs. Chocollama-8B-instruct | −0.004 | 0.001 | significant |
| Pattern vs. Chocollama-8B-instruct | −0.046 | 0.940 | non-significant |

### A.5.3 Distributional Analysis of the Models' Ratings

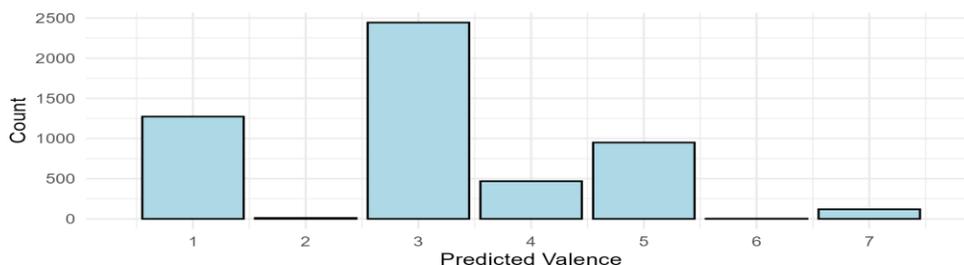

Figure 13: Distribution of Chocollama-8B-instruct's predicted valence scores

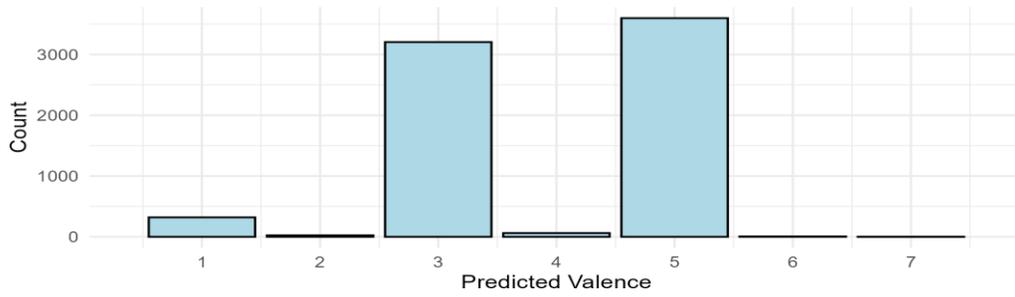

Figure 14: Distribution of Reynaerde-7B's valence scores

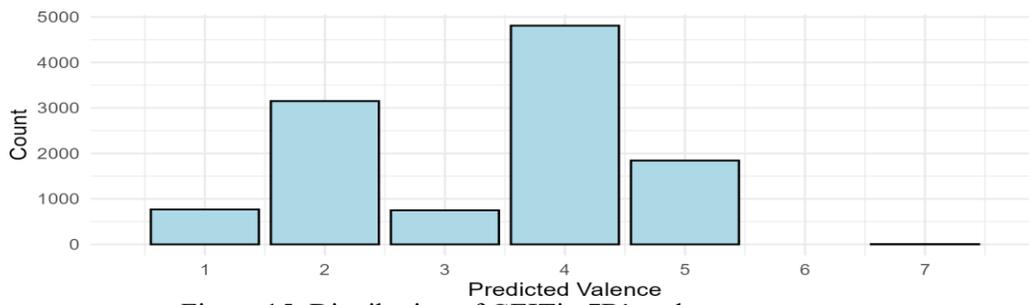

Figure 15: Distribution of GEITje-7B's valence scores

### A.5.4 Box Plots of Users' Ratings vs. Model's Predictions

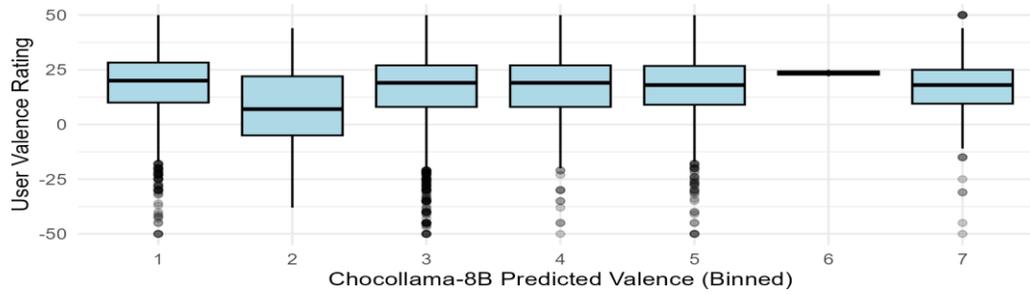

Figure 16: Chocollama-8B-instruct

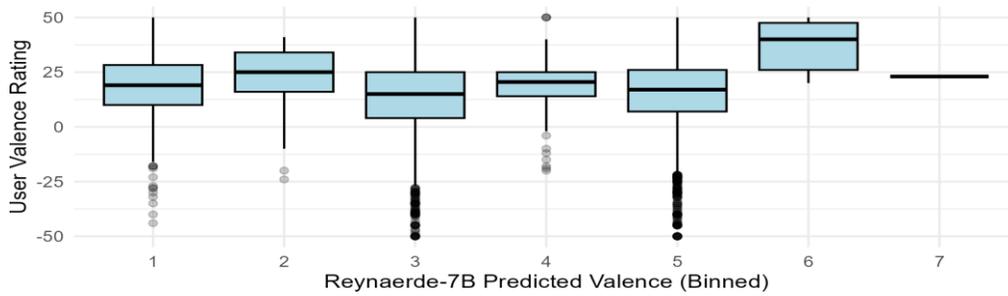

Figure 17: Reynaerde-7B

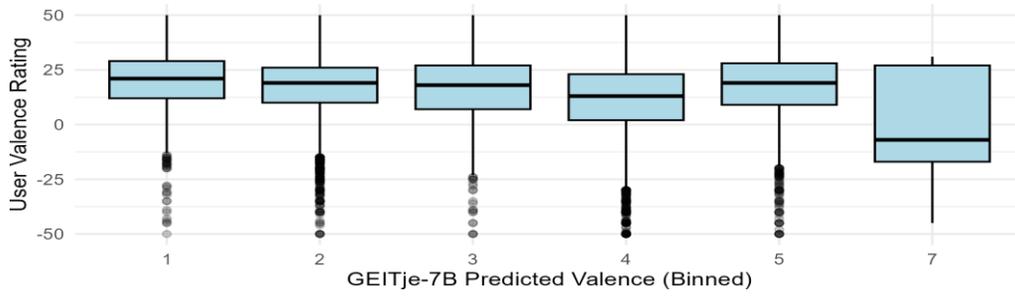

Figure 18: GEITje-7B

### A.5.5 Scatter Plots

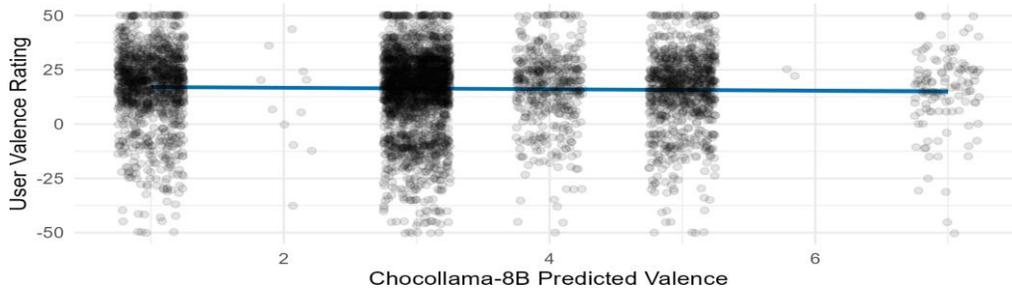

Figure 19: Chocollama-8B-instruct

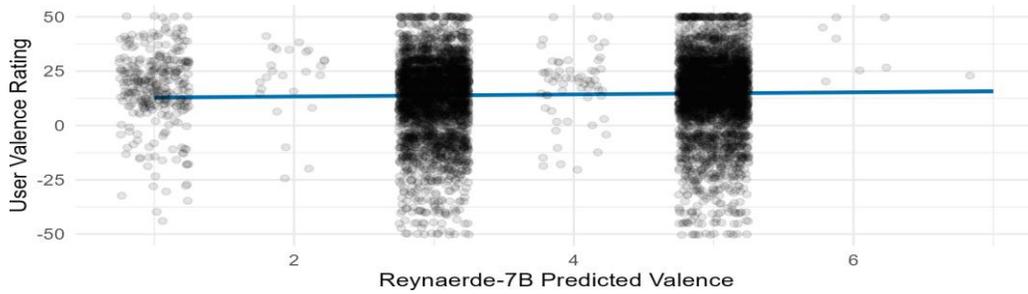

Figure 20: Reynaerde-7B

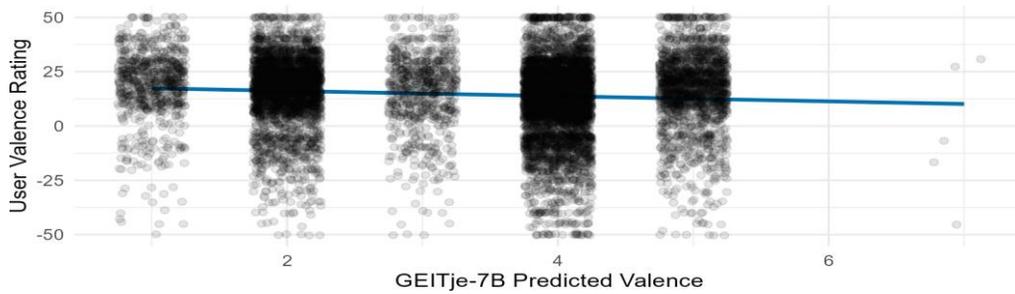

Figure 21: GEITje-7B

**A.6 References**